%% file: main.tex
\documentclass[letterpaper, 10 pt, conference]{ieeeconf}
\IEEEoverridecommandlockouts

\let\proof\relax
\let\endproof\relax
\usepackage{amsthm}
\usepackage[utf8]{inputenc}
\usepackage{amsmath,amssymb}
\usepackage{booktabs}
\usepackage{graphicx}
\usepackage{graphicx}
\usepackage{subcaption}
\usepackage[dvipsnames]{xcolor}
\usepackage{numprint}
\usepackage{csvsimple}
\usepackage{comment}
\usepackage[linesnumbered,ruled,noend]{algorithm2e}

\usepackage{tabularx,siunitx}
\usepackage{cite}
\usepackage{rotating} 
\usepackage{hyperref}
\usepackage[pscoord]{eso-pic} 

\newtheorem{theorem}{Theorem}
\newtheorem{problem}{Problem}

\newtheorem{lemma}{Lemma}

\newenvironment{proofsketch}{%
  \proof}{\endproof}

\newcommand{\norm}[1]{\left\lVert#1\right\rVert}
\newcommand{\reals}{\mathbb{R}}
\newcommand{\N}{\mathcal{N}}
\newcommand{\B}{\mathcal{B}}
\newcommand{\xhat}{\hat{x}}
\newcommand{\cov}{\Sigma_{k}}
\newcommand{\checkx}{\check{x}}
\newcommand{\checku}{\check{u}}
\newcommand{\checkX}{\check{X}}
\newcommand{\checkU}{\check{U}}

\newcommand{\X}{\mathcal{X}}
\renewcommand{\delta}{P_\text{safe}}

\newcommand{\placetextbox}[3]{
  \setbox0=\hbox{#3}
  \AddToShipoutPictureFG*{
    \put(\LenToUnit{#1\paperwidth},\LenToUnit{#2\paperheight}){\vtop{{\null}\makebox[0pt][c]{#3}}}%
  }%
}%

\title{\LARGE \bf Gaussian Belief Trees for Chance Constrained Asymptotically Optimal Motion Planning
}
\author{Qi Heng Ho, Zachary N. Sunberg, and Morteza Lahijanian
\thanks{This work was supported by the University of Colorado Boulder.}
\thanks{Authors are with the department of Aerospace Engineering Sciences at the University of Colorado Boulder, CO, USA
        {\tt\small \{\textit{firstname}.\textit{lastname}\}@colorado.edu}}%
}

\begin{document}
\placetextbox{0.5}{0.95}{To appear in IEEE International Conference on Robotics and Automation (ICRA), May. 2022.}
\maketitle
\input{sections/Abstract}
\input{sections/Introduction.tex}
\input{sections/Problem.tex}
\input{sections/Background.tex}
\input{sections/Methodology.tex}
\input{sections/extended/Analysis}
\input{sections/Experiments.tex}
\input{sections/Conclusion.tex}

\bibliographystyle{IEEEtran}
\bibliography{bib,zachbib}

\clearpage
\input{sections/Appendix.tex}

\end{document}

%% file: sections/Abstract.tex
\begin{abstract}
    
    In this paper, we address the problem of sampling-based motion planning under motion and measurement uncertainty with probabilistic guarantees. We generalize traditional sampling-based, tree-based motion planning algorithms for deterministic systems and propose belief-$\mathcal{A}$, a framework that extends any kinodynamical tree-based planner to the belief space for linear (or linearizable) systems. We introduce appropriate sampling techniques and distance metrics for the belief space that preserve the probabilistic completeness and asymptotic optimality properties of the underlying planner. We demonstrate the efficacy of our approach for finding safe low-cost paths efficiently and asymptotically optimally in simulation, for both holonomic and non-holonomic systems.
\end{abstract}

 

%% file: sections/Introduction.tex
\section{Introduction}

In recent years, sampling-based algorithms (e.g., \cite{kavraki1996prm, lavelle1998rrt, est1997, karaman2011sampling, cortes2020samplingbased}) have emerged as powerful motion planning tools as they can search high dimensional spaces very efficiently and find solutions to complex problems quickly. 
They have enabled many applications such as self-driving cars, surgical robots, and autonomous UAVs.
The majority of such planners assume perfect state information. 
However, an inseparable aspect of robotics is uncertainty in motion (e.g., due to modeling inaccuracies) and observation (e.g., due to noisy sensors).
This gap is especially significant when measurement uncertainty makes it difficult to find efficient paths while obeying  safety constraints.
This work focuses on this gap and aims to develop a general framework for adapting existing sampling-based algorithms to plan for uncertain systems with safety and optimality guarantees.



Consider a UAV equipped with a GPS receiver navigating under a canopy in a forest for a search and rescue task. The ability of the UAV to estimate its state is dependent on where it is in the forest, as some regions may have improved GPS accuracy or connectivity than others. To plan safe paths under dynamics and sensor noise, the robot has to trade off between visiting such regions or taking the shortest path to the goal location. Explicitly and intelligently accounting for both motion and measurement uncertainty is necessary to improve the quality and safety of motion plans.

In its most general form, planning under uncertainty can be formulated as a Partially Observable Markov Decision Process (POMDP) \cite{kaelbling1998planning}.
POMDP solvers
attempt to find an optimal policy in the belief space, which is the space of all possible probability distributions over the state space.
Unfortunately, this problem is computationally intractable due to the curses of dimensionality and history~\cite{papadimitriou1987complexity}.
Methods for approximating POMDP solutions for motion planning over continuous state, control, and observation spaces have been proposed, e.g., \cite{sunberg2018pomcpow,garg2019despotalpha}. 
However, they do not perform well especially when the action (control) space is large, which is a characteristic of robotic systems.  Another drawback is the lack of guarantees in these solutions.


In recent years, sampling-based algorithms have been extended to account for motion uncertainty with probabilistic guarantees, e.g., \cite{cc-rrt,blackmore2011, Pairet:TASE:2021, Hahn2019imdp, burgard2008smr, morteza-old, Luna:AAAI:2014, Luna:ICRA:2014, Luna:WAFR:2015}.  
Specifically, the chance-constrained tree-based planners have shown to be efficient, and hence, employed in  unknown environments via iterative planning~\cite{cc-rrt,blackmore2011, Pairet:TASE:2021, morteza-old}. A few studies extend those frameworks to systems with measurement uncertainty \cite{agha2014firm, Shan2017brm, Prentice2019brm, platt2010MLE}. 
For instance, in \cite{platt2010MLE, Prentice2019brm}, maximum-likelihood observations are used to approximate solutions.  While that approach works well in many cases, it lacks safety guarantees.
Work \cite{bry2011rrbt} proposes an alternative approach that provides chance-constrained guarantees by
combining optimal control and state estimation with sampling-based graphs. 
The method finds an optimal trajectory through the belief space
by constructing a graph of trajectories in the state space and enumerating all possible uncertainty levels for these trajectories in the belief space. 
That planner is asymptotically optimal and has probabilistic safety guarantees, but it suffers from large computation times. 

To mitigate that issue,
work \cite{Yang2016anytimerrbt} extended the method by using branch-and-bound pruning, which leads to linear computation speedups. However, it also modifies the cost function of the problem to include uncertainty, which may affect the optimality of its solutions with respect to the original cost function.



In this paper, we generalize tree-based motion planning to obtain a belief planning framework for linear (or linearizable) systems with both motion and observation uncertainty. This framework, called belief-$\mathcal{A}$, enables the use of any kinodynamical planner $\mathcal{A}$ for belief space motion planning. Belief-$\mathcal{A}$ provides chance constraint guarantees, and preserves the probabilistic completeness and asymptotic optimality properties of the underlying planner as well as its computational efficiency.
This is achieved by utilizing appropriate sampling techniques and distance metrics in the belief space.

Specifically, our contributions are (i) a general method to extend state space sampling-based tree search algorithms directly to the Gaussian belief space, (ii) a method to uniformly sample in the belief space, (iii) a way to heuristically bias sampling towards low uncertainty belief states, and (iv) the use of the 2-Wasserstein metric for distance computation between beliefs. 
We show the efficacy of our framework through benchmarking and numerical characterization in several scenarios that highlight the relative strengths of different aspects of our framework. The results indicate that our method performs much faster (up to 50 times speedup) compared to prior belief space tree search methods while still having asymptotically near-optimal guarantees, allowing for safe and efficient planning. As a by-product of the computational speedups, we also show the ability of our planner to conduct online re-planning when knowledge of the environment changes.


%% file: sections/Problem.tex
\section{Problem Formulation}
\label{sec:problemstatement}

We are interested in motion planning for a robotic system with uncertainty in both motion and observation (measurement) with safety guarantees. 
We assume the robot evolves in a bounded workspace ($\reals^2$ or $\reals^3$) with obstacles according to linear or linearizable dynamics and a measurement model that can change in different parts of the environment, e.g., GPS signal may be available only in parts of the environment.  Below, we formalize this problem.

Let $\mathcal{X} \subset \reals^n$ and $\mathcal{U} \subset \reals^m$ be the state and control spaces, respectively.
Then, the robot model is given by:
\begin{equation}
    \label{eq:system}
    \begin{split}
        x_k &= Ax_{k-1} + Bu_{k-1} + w_k, \quad w_k \sim \N(0, Q),\\
        z_k &= Cx_k + v_k(x_k), \quad v_k(x_k) \sim \N(0, R(x_k)),
    \end{split}
\end{equation}
where $x_k \in \mathcal{X}$ is the state, $u_k \in \mathcal{U}$ is the input, $z_k \in \reals^p$ is the measurement.
Furthermore, matrices $A \in \reals^{n\times n}$, $B \in \reals^{n\times m}$, and $C \in \reals^{n\times p}$.
Terms $w_k$ and $v_k$ are i.i.d white Gaussian noise with zero mean and $Q$ and $R(x_k)$ covariance matrices, respectively. 
Note that a unique aspect of this robot model is that
the measurement noise (covariance matrix) is a function of the state, representing, e.g., various types of measurement regions in the environment.

We assume the system's dynamics are fully controllable and observable, and the noise covariance matrices $Q$ and $R(x_k)$ are non-degenerate. 
For presentation simplicity, 
we focus on the time-invariant system in \eqref{eq:system}, 
but the formulation can be easily extended to time-varying systems.  

The evolution of the robotic system in \eqref{eq:system} can be described by a discrete-time Gaussian Markov process. The robot state $x_k$ at each time step can be described as a Gaussian distribution $b_k = \N(\xhat_k, \cov)$, where $b_k \in \B$ is referred to as the belief of the robot state, $\B$ is the \textit{belief space} of the robot, and $\xhat_k$ and $\cov$ are the state mean and covariance matrix, respectively.  
In this work, we search for a motion plan characterized by a sequence of nominal control inputs and the nominal trajectory that they produce in the absence of uncertainty, along with a linear feedback controller that seeks to follow this trajectory using online state estimate $\xhat_k$.

Let $\checkU^{0,t} = (\checku_0, \checku_1, \ldots, \checku_{t-1})$ be a sequence of control inputs. Given an 
initial state $x_0$ and nominal system dynamics $\checkx_{k+1} = A\checkx_{k} + B\checku_k$, a nominal trajectory
$\checkX^{x_0,x_t} = (\checkx_0, \checkx_1, \ldots, \checkx_t)$ is obtained. This motion plan is then executed online via a stabilizing controller 
given online state estimates $\xhat_k$. This gives us the following online control input
\begin{align}
    \label{eq:feedback controller}
    u_k = \check{u}_{k-1} - K (\xhat_k - \checkx_k),
\end{align}
where $K$ is the closed loop gain. For simplicity, we use a fixed controller gain, but a variable gain can also be used. 
We refer to $(\checkU,\checkX)$ as a \textit{motion plan} for System \eqref{eq:system}.




Let $\mathcal{X}_{obs}, \mathcal{X}_{goal} \subset \mathcal{X}$ represent the 
state space obstacles and the goal region 
respectively. 
For robot state $x_k$ and its belief $b_k$, the probability of 
the robot being in $\X_i$, where $i \in \{{obs}, {goal}\}$, is given by
\begin{align}
    \label{eq:probabilityinregion}
    p(x_k \in \X_i) = \int_{\X_i} b_k(y) dy.
\end{align}
Furthermore, consider a stage cost function $J: \X \to \reals^+$ that is monotonic and Lipschitz continuous.  
Given safety probability $\delta$,
we desire an optimal motion plan that minimizes total cost given by $J$ and whose probability of collision at every time step and probability of not ending in goal are upper bounded by $\delta$. 
The formal statement of this problem is as follows.

\begin{problem}
    \label{env:problem}
    Given a robot with noisy dynamics and measurements as in \eqref{eq:system}, a set of state space obstacles $\X_{obs}$, a goal region $\X_{goal}$, a Lipschitz continuous cost function $J: \X \to \reals^+$, and a safety probability bound $\delta$, find a motion plan $(\checkU, \checkX)^*$ as a pair of sequence of nominal controls $\checkU = (\checku_0,\ldots,\checku_{T-1})$ for some $T\geq 1$ and its resulting nominal trajectory $\checkX = (\checkx_0,\ldots, \checkx_T)$ that minimizes the expected total cost, 
    \begin{equation}
        (\checkU, \checkX)^* = \arg\min_{(\checkU, \checkX)}\, \mathbb{E} \Big[ \sum_{t=0}^{T} J(x_{t}) \Big],
    \end{equation}
    subject to, when executed via controller in \eqref{eq:feedback controller}, the probabilities of collision with $\X_{obs}$ and not ending in $\X_{goal}$ are bounded by $\delta$, i.e.,
    \begin{align}
        &p(x_t \in \mathcal{X}_{obs}) < \delta ,\quad \forall t \in [0, T],\\
        &p(x_T \notin \mathcal{X}_{goal}) < \delta.
\end{align}
    
\end{problem}


To solve this optimal motion planning problem, we aim to design an algorithm that is asymptotically optimal, i.e. the probability of finding the optimal solution converges to 1 as the number of iterations approaches infinity. For our applications of interest, asymptotic near-optimality (i.e., the solution converges to within an approximation factor $\epsilon$ of the optimal solution) is sufficient, but our proposed framework is not limited to asymptotic near-optimality.


%% file: sections/Background.tex
\section{Preliminaries}
Here, we introduce the preliminaries required to introduce our solution to the above problem.

\subsection{Asymptotically-Optimal Tree Search Planners}
\label{sec:treebasedplanners}
Typical single query sampling-based tree search algorithms construct a tree in the search space. For kinodynamical systems without any uncertainty, this search space is the state space. In this work, we aim to develop a method to transform such existing planners into belief space planners that solve Problem~\ref{env:problem}. To this end, we first present an overview of such planners.


Algorithm~\ref{alg:treebasedplanners} shows a generic form of an asymptotically optimal tree-based planner for kinodynamical systems.
It takes state space $\mathcal{X}$, input space $\mathcal{U}$, goal region $\mathcal{X}_{goal}$, obstacle regions $\mathcal{X}_{obs}$, an initial state $x_{init}$, and a maximum planning time or iteration count (N) as input and returns a near-optimal solution, if one is found. Search is performed by growing a motion tree in which states and the connections between them are stored as nodes in $\mathbb{V}$ and edges in $\mathbb{E}$, respectively. Note that the Prune subroutine is not mandatory for asymptotic optimality, but many asymptotically optimal planners, e.g., the Stable Sparse-RRT (SST) \cite{SST}, utilize it.

\begin{algorithm}
    \SetKwInOut{Input}{Input}
    \SetKwInOut{Output}{Output}
    \caption{Generic Asymptotically-Optimal Tree-based Planner $\mathcal{A}$ ($\mathcal{X}, \mathcal{U}, \mathcal{X}_{goal}, \mathcal{X}_{obs}, x_{init}$, N)}
    \label{alg:treebasedplanners}
    \Output{Valid Trajectory $x_{1:T}$ if one is found}
    $G = (\mathbb{V} \leftarrow \{x_{init}$\}, $\mathbb{E} \leftarrow \emptyset)$\\
    \For{N iterations}{
        $x_{rand} \leftarrow$ Sample()\\
        $n_{select} \leftarrow$ Select($x_{rand}$)\\
        $n_{new} \leftarrow$ Extend($n_{selected}$)\\
        \uIf {isValidPath($n_{select}, n_{new}$)}{
            $\mathbb{V} \leftarrow \mathbb{V} \cup \{n_{new}$\}\\
            $\mathbb{E} \leftarrow \mathbb{E} \cup \{edge(n_{select}, n_{new})\}$\\
        }
        Prune($\mathbb{V}, \mathbb{E}$)
    }
    \Return $G=(\mathbb{V}, \mathbb{E})$
\end{algorithm}
\vspace{-1em}

\subsection{Propagation of Beliefs through Feedback Dynamics}
\label{sec:propagation}
Following \cite{bry2011rrbt}, given a stabilizing feedback controller in \eqref{eq:feedback controller} and online state estimate $\xhat_k$, the system closed-loop dynamics are
\begin{align}
    x_k = Ax_k + B(\check{u}_{k-1} - K(\hat{x}_k - \check{x}_k)) + w_k\text{.}
\end{align}
The belief during planning is parameterized as
\begin{align}
    \label{eq:belief}
    b_k = P(x_k) = \N(\checkx_k, \cov^+ + \Lambda_k^+ ),
\end{align}
where $\cov^+$ is the covariance that represents the online state estimation error, and $\Lambda_k^+$ is the uncertainty over possible state estimates that could arise during execution and is a result of considering all possible measurements received.
$\cov^+$ and $\Lambda_k^+$ can be computed recursively using the Kalman Filter:
\begin{align}
    \cov^- &= A\Sigma_{k-1}^{+}A^T + Q,\\
    L_k &= \cov^-C^T(C\cov^-C^T + R(\xhat_k))^{-1},\\
    \cov^+ &= \cov^- - L_kC\cov^-,\\
    \Lambda_k^+ &= (A- BK)\Lambda_{k-1}^+(A-BK)^T + L_kC\cov^- \text{.}\label{eq:beliefpropagation}
\end{align}

%% file: sections/Methodology.tex
\section{Gaussian Belief Trees}

Here, we present our methodology for finding chance-constrained motion plans in the belief space to solve Problem~\ref{env:problem}.
We present our framework in its general form, which can be used with any single-query state space tree-based motion planner with a structure similar to Algorithm~\ref{alg:treebasedplanners}. 
Instead of planning in the state space, we plan in the belief space $\B$, where $b \in \B$ is a Gaussian distribution as defined in \eqref{eq:belief}.
Vertices in the tree structure are now belief nodes, rather than state nodes. 

As discussed in Section \ref{sec:treebasedplanners} and Algorithm \ref{alg:treebasedplanners}, the main subroutines for tree search algorithms are the sample, select, extend, and validity checking.
Below, we present methods for reformulating these functions for reasoning in the belief space.
Specifically, we show how to sample directly in the belief space, and propose a method for heuristically biasing the sampling. Then, we present a belief space distance function for node selection through the use of a suitable distance metric for probability distributions. Finally, we present our approach to belief propagation and validity checking.
\subsection{Sampling Strategy}
\label{sec:sampling}
Since our beliefs are Gaussian distributions, the Sample function must randomly sample the first moment (mean $\mu_s$)
and second moment (covariance matrix $\Sigma_s$).

\subsubsection{Sampling of first moment}

Similar to state space planning, we sample the mean of a belief from the state space, such that $\mu_s \in \mathcal{X}$. We also utilize a goal bias, in which we sample $\mu_s$ within the goal region with a probability $p_{goal}$, which yields faster convergence in practice\cite{Urmson2003biasrrt}.

\subsubsection{Sampling of covariance matrices}

The optimal sampling strategy for motion planning in Gaussian belief spaces is an open question, but it is crucial that the method for sampling covariance matrices completely samples the space of covariances for a problem environment in order to preserve probabilistic completeness of the state space version of the algorithm. We show how to uniformly sample the space of covariance matrices for a given environment.

We first observe that since Q and R are assumed to be non-degenerate, the covariance matrix of $b_k$ is necessarily positive definite.
Hence it is always diagonalizable, i.e., we can write $\Sigma_s = O D O^{T}$, where $D$ is a $n \times n$ diagonal matrix whose diagonal elements consist of its non-negative eigenvalues and $O$ is a $n\times n$ orthogonal matrix whose columns consist of its linearly independent real and orthonormal eigenvectors. Therefore, we can sample Gaussian covariance matrices by generating eigenvalues for $D$ and an orthogonal matrix $O$ of eigenvectors. 
To do this, we first uniformly sample eigenvalues $\lambda_i$ from the interval $(0, \lambda_{i,max}]$ where $\lambda_{i,max}$ defines the maximum allowed eigenvalue for the $i^\text{th}$ dimension. $\lambda_{i,max}$ depends on the environment. One method for setting $\lambda_{i,max}$ is to find the minimum $\lambda_{i} > \lambda_{i,max}$ such that forming a diagonal covariance matrix with $\lambda_i$ as the $i^\text{th}$ diagonal element and setting the rest of the diagonal elements as a small $\epsilon$ violates the chance constraint at every state in the state space. 

Next, to uniformly sample random orthogonal matrices $O$, we first generate an $n\times n$ matrix $M$ whose elements are independently sampled from a standard normal distribution (with mean 0 and standard deviation 1).
The QR decomposition of $M$, i.e., $M = OR$, then provides a uniform distribution of an orthogonal matrix $O$ \cite{Eaton1985MultivariateS}\footnote{Sampling elements of $M$ with another distribution, e.g. uniform, still randomly samples an orthogonal matrix $O$, but not necessarily uniformly.}. Finally, we obtain our randomly sampled Gaussian covariance matrix by computing $\Sigma_s = ODO^{T}$.

\subsubsection{Sampling bias}

To improve performance, we can exploit the intuition that low uncertainty beliefs result in reduced collision probability, which improves the chances of finding new valid edges and nodes. Additionally, since the problems we are interested in have an inherent tradeoff between motion plans that have lower cost and those that have higher cost but lead to lower uncertainty trajectories, this low uncertainty bias provides a way to prioritize lower uncertainty paths without explicitly considering it in the cost.

Therefore, with probability $p_{bias}$, we sample a fixed low eigenvalue $\lambda_{i,low}$ instead of uniformly in  the interval $(0, \lambda_{i,max}]$. This effectively pulls the tree towards low uncertainty areas in the belief space, behavior empirically shown in our experiments to be advantageous, while still ensuring completeness. The optimum $p_{bias}$ depends on the environment, where a high $p_{bias}$ is useful for environments in which solution paths require low uncertainty but may slow down convergence if low uncertainty is not required. In the general case, $p_{bias}$ should be set to a low value to ensure a more general coverage of the search space.
\subsection{Belief Space Metric}
A key component of many tree-based algorithms, such as RRT~\cite{lavelle1998rrt} and SST~\cite{SST}, is a distance metric to compute distances between nodes in the tree. This metric is used for near neighbor computations, to select which node to extend in the Select function, and for pruning. A good choice of distance metric can greatly affect exploration of the search space, planning time and path quality \cite{Palmieri2015DistanceML, littlefield2018}. 

In the Gaussian belief space, a typical state space distance metric such as Euclidean distance can still be used on the means of the belief nodes, but that is not ideal as they do not capture the uncertainty of the belief state. Instead, we use the Wasserstein distance \cite{panaretos2019wasserstein}, which is a metric on a space of probability measures. Intuitively, the Wasserstein metric is the amount of work required to move the probability mass of one distribution to another. For two non-degenerate Gaussian beliefs $b_1$ and $b_2$ with $b_i \sim N(\mu_i, \Sigma_i)$, with respect to the Euclidean norm on $\reals^n$, the 2-Wasserstein distance is
\begin{multline}
    \label{eq:wasserstein}
    D_{W_2}(b_1, b_2) = \\
    \norm{\mu_1 - \mu_2}^2_2 + Tr(\Sigma_1 + \Sigma_2 - 2(\Sigma_1^\frac{1}{2}\Sigma_2\Sigma_1^\frac{1}{2})^\frac{1}{2}),
\end{multline}
where $Tr(M)$ is the trace of $M$. Unlike other belief distance functions such as KL-divergence and LP-distance, the Wasserstein metric is a true metric in the belief space and has desirable properties for the preservation of asymptotic optimality of the state space algorithm (see Section~\ref{sec:convergence}).

\subsection{Extend/Propagate}

After selecting a belief node, using the sampling method and distance metric described above, kinodynamic tree-search algorithms create a new node in the tree by extending the selected node using a computed nominal control input and random time duration. 
Given a belief node $b$, nominal control inputs $\checku$ and a time duration, we propagate the belief to a new belief node $b'$ using \eqref{eq:belief}-\eqref{eq:beliefpropagation}. The control inputs can be computed by either repeatedly sampling the control space directly or sampling a state $x_{sample}$ and applying a closed loop controller for a sampled time duration. 

It is important to note that although the system is linear and controllable which implies that we can always compute a control input to steer a state node to any other state, steering between two Gaussian belief nodes involves not only steering the means from one belief to another, but propagation from one covariance to another. Even for a linear system, a solution for controlling an arbitrary Gaussian covariance to converge to another arbitrary covariance is not guaranteed to exist. Therefore, tree-search algorithms that rely on a steering function between two nodes, such as RRT$^*$, cannot be readily extended to the belief space via our method.


\subsection{Validity Checking}

Validity checking in the state space involves deterministic collision checking, i.e. checking if  $x \in \mathcal{X}_{obs}$.
In the belief space, it involves computing the probability of collision $p(x) \in \mathcal{X}_{obs}$ in \eqref{eq:probabilityinregion}, which involves an integral. The exact evaluation of this integral over obstacles is not always computable, and numerical integration is computationally expensive. Efficient conservative approximations can be computed very quickly for convex polygonal obstacles using error functions as proposed in \cite{blackmore2011}. For non-convex obstacles, methods such as \cite{park2017, Pairet:TASE:2021} provide high accuracy at the cost of computation time.





%% file: sections/extended/Analysis.tex
\section{Analysis}

\subsection{Completeness and Optimality}
\label{sec:convergence}


The belief space extension inherits the probabilistic completeness and asymptotic (near)-optimality properties of the state space versions of the algorithms. The following theorems formalize these properties.

\begin{theorem}[Completeness]
    Algorithm belief-$\mathcal{A}$ is probabilistically complete for the stochastic system in \eqref{eq:system}
    if $\mathcal{A}$ is probabilistically complete for the nominal system in \eqref{eq:system} (without stochasticity).
\end{theorem}

\begin{proofsketch}
    Probabilistic completeness for tree-search algorithms requires that the search space is completely sampled almost surely.
    The main relevant consequence of the change to the belief space is the difference in sampling of the search space and propagation dynamics. However, with the sampling method described in~\ref{sec:sampling}, beliefs are sampled uniformly, which implies the search space is completely sampled almost surely (the inclusion of the bias term does not change the result). Additionally, if the dynamics is Lipschitz continuous, the propagation method presented in the belief space is also Lipschitz continuous. Hence, Belief-$\mathcal{A}$ inherits the probabilistic completeness properties of $\mathcal{A}$. 
\end{proofsketch}

The use of the 2-Wasserstein metric makes the new belief space a proper metric space, and it is known that the 2-Wasserstein metric preserves the Lipschitz continuity of the cost function~\cite{littlefield2018,kurniawati2013}. 
\begin{lemma}Theorem in \cite{littlefield2018}.
    \label{lemma:lipschitz}
    For any control input $u \in \mathcal{U}$, if the cost function in the state space is Lipschitz continuous in the state space, then the expected cost function in the belief space with the 2-Wasserstein metric is also Lipschitz continuous.
\end{lemma}

\noindent Convergence guarantees for asymptotically ($\epsilon$-)optimal sampling-based planners typically rely on Lipschitz continuity in the cost function \cite{littlefield2018}.
This motivates the following theorem:

\begin{theorem}[Optimality]
    Belief-$\mathcal{A}$ is asymptotically ($\epsilon$-)optimal if algorithm $\mathcal{A}$ is asymptotically ($\epsilon$-)optimal under the conditions in Problem~\ref{env:problem}.
\end{theorem}
\begin{proofsketch}
    Since the cost function in Problem~\ref{env:problem} is Lipschitz continuous in the state and control space, by Lemma~\ref{lemma:lipschitz}, the cost function in the belief space under the 2-Wasserstein metric is also Lipschitz continuous. For belief-$\mathcal{A}$, the control space, method to sample controls and node selection procedure remains the same as $\mathcal{A}$. Hence, Belief-$\mathcal{A}$ inherits the asymptotic ($\epsilon$-)optimality properties of $\mathcal{A}$.
\end{proofsketch}
\subsection{Computational Efficiency}

The overall time complexity of the underlying tree search algorithm does not change. However, planning in the belief space increases the computational effort in two ways. Firstly, planning in the belief space changes the dimension of the search space changes the dimension of the search space. Additionally, our extensions change specific subroutines, giving a computation time increase per call.

\subsubsection{Search Dimension}

Planning in the belief space changes the dimension of the search space from $n$ state dimensions to $n + n^2$ Gaussian belief dimensions, which leads to a higher dimensionality of search space increases planning time. However, sampling-based tree motion planners scale very well to moderately high dimensions since a relatively sparse set of sampled nodes may be necessary to construct a valid solution plan.

\subsubsection{Distance Metric}

The use of the Wasserstein metric for nearest neighbor queries is slower than usual state space metrics. For general belief distributions, the Wasserstein metric has a time complexity of $O(N^3log N)$ for $N$ bin histograms. However, since our beliefs are Gaussian distributions, the Wasserstein distance computation in \eqref{eq:wasserstein} is more efficient, $O(n^3)$. This is higher than typical state space distance computations of $O(n)$. However, since $n$ is fixed and does not grow with the number of iterations, this leads to a bounded computational time increase per iteration.

\subsubsection{Sampling in the Belief Space}

Similarly, the time complexity of sampling random $n\times n$ matrix is $O(n)$, and that of QR decomposition is $O(n^3)$. Therefore, we have an overall time complexity for sampling of $O(n^3)$, which is higher than state space sampling of $O(n)$. Although this is more computationally expensive than usual state space sampling, since n is fixed and does not grow with the number of iterations, this also leads to a bounded computational time increase per iteration.

\subsubsection{Node Propagation}

Node propagation using the method described in \ref{sec:propagation} involves simple matrix additions, multiplications, and inversions. Matrix inversions have a time complexity of $O(n^{2.373})$, which is the dominant term in the propagation routine. This is similar to the state space extension routine, of which time complexity is dominated by matrix multiplications with a time complexity of $O(n^{2.372})$. However, there are more operations per node propagation in the belief space than for state space planning, leading to a higher computational cost per call.


\subsubsection{Validity Checking}
In the belief space, the computational cost of the validity checking subroutine depends on the method used. If the obstacles are convex, using the validity checking method described in \cite{blackmore2011}, the approximation of integrals for~\eqref{eq:probabilityinregion} can be computed as deterministic linear constraints very efficiently, albeit with some conservatism.

Empirically, as seen in the experiments, the belief space extension subroutines are reasonably efficient, allowing for fast planning in the belief space. We also compared to other approaches in the literature which solve similar problems in the belief space, in which our method is shown to have superior computational efficiency.



%% file: sections/Experiments.tex
\section{Experiments and Results}

\subsection{Time and Path Length Benchmarks}
We compare our methodology on two systems, a 2D system with dynamics
\begin{align*}
    x_k = x_{k-1} + u_{k-1} + w_k, \quad
    z_k = x_k + v_k(x_k),
\end{align*}
and a second order unicycle system with dynamics
\begin{align*}
    \dot{x} = v \cos{(\phi)}, \quad
    \dot{y} = v \sin{(\phi)}, \quad \dot{\phi} = \omega,  \quad \dot{v} = a\text{.}
\end{align*}
We use dynamic feedback linearization to obtain a linearized closed loop model as presented in \cite{DELUCA2000}.

\begin{figure}
        \centering
        \begin{subfigure}[b]{0.15\textwidth}
            \centering
            \includegraphics[width=\textwidth]{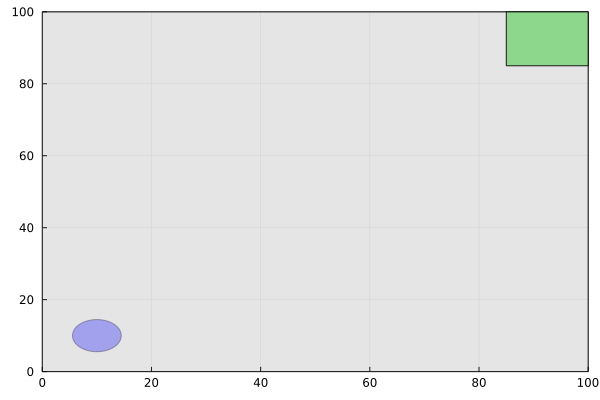}
            \caption[Environment 1]%
            {{\small Environment 1}}    
            \label{fig:initial1}
        \end{subfigure}
        \begin{subfigure}[b]{0.15\textwidth}  
            \centering 
            \includegraphics[width=\textwidth]{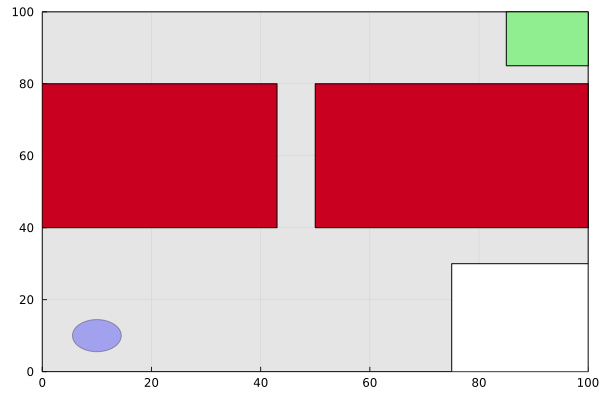}
            \caption[Environment 2]%
            {{\small Environment 2}}    
            \label{fig:replan1}
        \end{subfigure}
        \begin{subfigure}[b]{0.15\textwidth}   
            \centering 
            \includegraphics[width=\textwidth]{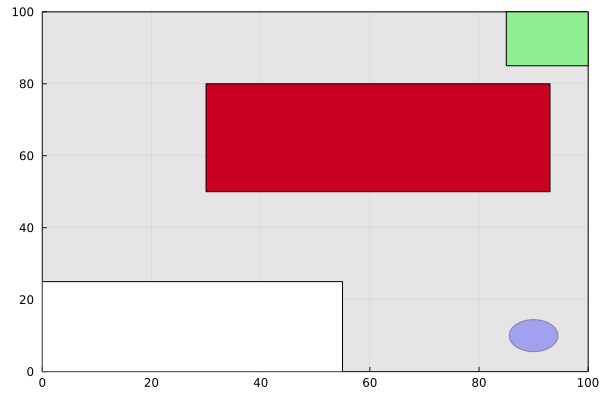}
            \caption[Environment 3]%
            {{\small Environment 3}}    
            \label{fig:replan1sec1}
        \end{subfigure}
        \caption[]
        {{\small The initial belief, goal region and obstacles are represented as blue, green and red respectively. White region are measurement regions, in which the measurement noise covariance $R(x_k) = 0.01\mathit{I}$, and with no measurements everywhere else (gray). Environments are $100 \times 100 ~(m^2)$.}}
        \label{fig:environment}
        \vspace{-6mm}
\end{figure}

We conducted experiments in specific environments that force a trade-off between motion plans that visit measurement regions to localize and those that correspond to shorter path lengths. These environments, depicted in Fig.~\ref{fig:environment}, are
\begin{enumerate}
    \item Non-observable environment with no obstacles and no measurement region.
    \item Narrow passage environment with a measurement region. Moving straight to the goal will not satisfy the chance constraint.
    \item Multi mode solution environment. Two main solution paths are available - A larger passageway that allows reaching the goal region without measurements, and a narrow passage that requires a lower uncertainty.
\end{enumerate}

\begin{table*}[!ht]
    \centering
    \begin{tabular}{|c|c|c|c|c|c|c|c|c|c|}
    \hline
    & \multicolumn{3}{c|}{Environment 1} & \multicolumn{3}{c|}{Environment 2} & \multicolumn{3}{c|}{Environment 3}\\
    \hline
    Algorithm & Time (1st) & Cost (1st) & 
    Cost (10s) & Time (1st) & Cost (1st) & 
    Cost (10s) & Time (1st) & Cost (1st) & 
    Cost (10s)\\
    \hline
     \multicolumn{10}{|c|}{2D System}\\
    \hline
     RRBT   & 0.17 & \textbf{113.8}  & \textbf{109.8}  &  1.56  & \textbf{216.1}  & \textbf{199.2}  & 2.24  & \textbf{146.8}  & \textbf{121.2} \\\hline
     B-RRT ($l_2$)& \textbf{0.003} & 194.8   & 192.6  & \textbf{0.024}  & 360.8  & 341.7  & \textbf{0.009} & 293.0  & 292.6 \\\hline
     B-RRT ($W_2$)&  0.029  & 195.7  & 186.1  & 0.2  & 353.7  & 345.1  & 0.016  & 288.6  & 285.3  \\\hline
     B-RRT ($W_2$, $0.2\,p_{bias}$) & 0.029  & 189.8  & 185.8  & 0.071   & 348.4  & 340.0  & 0.016  & 294.9  & 293.8  \\\hline
     B-SST ($l_2$) & \textbf{0.004}  & 163.6  & \textbf{111.8}  & 2.62  & 259.4  & 236.1  & \textbf{0.008}  & 264.9  & 166.9 \\\hline
     B-SST ($W_2$) & 0.029  & 181.3   & \textbf{111.1}  & 0.51  & 295.2  & 236  &  0.043  & 263.9   & 157.4  \\\hline
     B-SST ($W_2$, $0.2\, p_{bias}$) & 0.027  & 168  & \textbf{110.2}  & 0.152 & 304.7  & \textbf{204.9}  &0.041 & 263.2  & \textbf{128.6}  \\\hline
      \multicolumn{10}{|c|}{Linearized Unicycle}\\
     \hline
     RRBT    & 0.225   & \textbf{123.9}  & \textbf{120}  & 15.4  & \textbf{227.1}  & \textbf{222.6}  & 8.95  & \textbf{201.6}  & 199.4  \\\hline
     
     B-RRT ($l_2$)& \textbf{0.011}  & 250.0  & 250.0  & \textbf{0.447}  & 399.4  &367.7   & 0.045  & 391.9  & 390.9 \\\hline
     
     B-RRT ($W_2$)& 0.034   & 247.6  & 221.0  & 0.577  & 286.4  & 256.3  & 0.2  & 341.8 &268.3 \\\hline
     
     B-RRT ($W_2$, $0.2\,p_{bias}$) & 0.041  & 244.8   & 231.6  & 0.61   & 287.0  & 260.8  & 0.19  & 332.5  & 289.3 \\\hline
     
     B-SST ($l_2$) & 0.013  & 198.7  & 138.5  & 7.08  & 267.4 & \textbf{237.9}  & \textbf{0.035}  & 339.3 & \textbf{177.8}  \\\hline
     
     B-SST ($W_2$) & 0.055  & 190.3  & 137.2  & 2.19  & 277.8  & \textbf{232.1}  & 0.193  & 347.2 & 188.7 \\\hline
     
     B-SST ($W_2$, $0.2\, p_{bias}$) & 0.054  & 193.8  & 140.4   &  0.88  & 274.5  & \textbf{231.1}  & 0.153  & 316.1  & \textbf{164.4} \\\hline
    \end{tabular}
    \caption{\small Benchmark planner performance results. Each entry is the mean of 100 simulations. Quantification of the standard error may be found in Appendix~\ref{appendix:benchmarks} and is generally significantly less than the variation between rows. The scores within 10\% of the best score in each column are bolded.}
    \label{tab:benchmarks}
    \vspace{-4mm}
\end{table*}

We compared the effect of using the Wasserstein metric ($W_2$) with the Euclidean metric ($l_2$), and including belief sampling bias $p_{bias}$ for belief-RRT (B-RRT) and belief-SST (B-SST). We also ran benchmarks against Rapidly-exploring Random Belief Trees (RRBT) \cite{bry2011rrbt}, a single query RRG-based optimal planner that solves similar problems. For each case, we use $\delta = 0.95$ and the RRT goal bias $p_{goal} = 0.05$.
For all experiments, the cost function is path length.
All algorithms are implemented on the Open Motion Planning Library (OMPL)~\cite{sucan2012the-open-motion-planning-library}, and all computations performed single-threaded on a nominally 2.20 GHz CPU.


 Table~\ref{tab:benchmarks} shows the time taken to compute a first solution, the cost of the first solution, and the solution cost after 10 seconds for each of the environments and dynamics. Belief-SST consistently finds initial solutions very quickly and given additional time, optimizes solutions to low path costs. Using the Euclidean metric instead of the Wasserstein metric computes each iteration faster, but does not find solutions quickly in narrow passageway situations due to the pruning subroutine causing lower uncertainty but higher cost paths to repeatedly be pruned. Belief-RRT finds solutions the fastest but does not improve solutions much when given more time, which is expected as RRT is not asymptotically optimal.
 
 Finally, RRBT computes high quality initial solutions but takes a very long time to do so. This is expected since RRBT exhaustively propagates candidate solutions in the belief space from a state space RRG, so solution paths once found already have low cost. However, RRBT maintains an excessive number of belief nodes for each state RRG vertex that get propagated to a new vertex at each iteration, immensely slowing computation.

\begin{figure}[ht]
    \centering
    \begin{subfigure}[b]{0.23\textwidth}
            \centering
            \includegraphics[width=\linewidth]{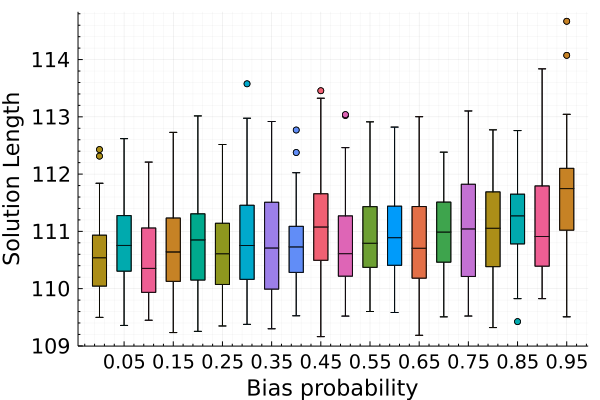}
            \caption[Environment 1.]%
            {{\footnotesize Environment 1.}}    
            \label{fig:biasenv1}
        \end{subfigure}
        \begin{subfigure}[b]{0.23\textwidth}  
            \centering 
            \includegraphics[width=\linewidth]{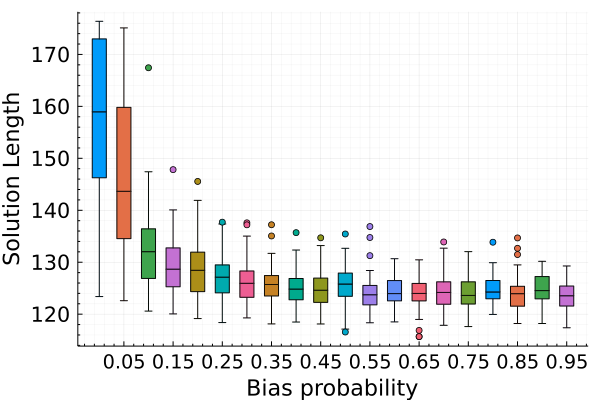}
            \caption[Environment 3.]%
            {{\footnotesize Environment 3.}}    
            \label{fig:biasenv3}
        \end{subfigure}
        \caption[Effect of varying $p_{bias}$ on solution length with 10 seconds of planning time.]
        {\small Effect of varying $p_{bias}$ on solution length with 10 seconds of planning time.
        }
        
        \label{fig:biaseffect}
        \vspace{-5.5mm}
\end{figure}


\subsection{Sampling Bias}

We study the effect of different $p_{bias}$ bias values for the sampling of low uncertainty covariances on solution length and time. The performance of varying bias values with belief-SST is depicted in Fig.~\ref{fig:biaseffect} (results for Environment 2 are similar to that of Environment 3). The bias has a direct effect on the rate of convergence to low solution costs. In Environment 3, a higher $p_{bias}$ leads to faster convergence to better solutions. However, there is a small trade-off in solution cost for Environment 1, in which low uncertainty is not needed. In general, a low value such as $p_{bias} = 0.2$ works well, leading to large gains for environments that require low uncertainty while also performing well in those that do not.

\subsection{Belief-SST with covariance sampling bias}

\begin{figure}
        \centering
        \begin{subfigure}[b]{0.23\textwidth}
            \centering
            \includegraphics[width=\textwidth]{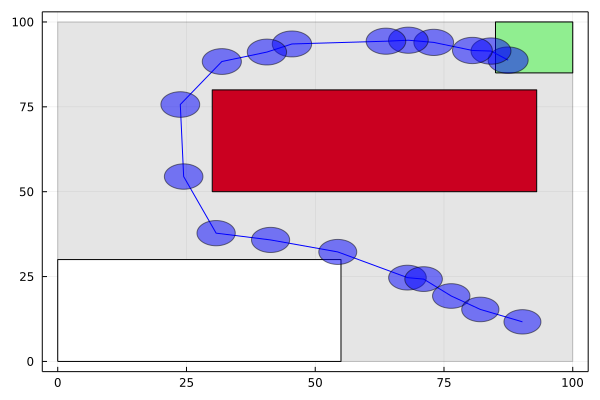}
            \caption[Initial plan]%
            {{\footnotesize 1 second planning time}}    
            \label{fig:initial2}
        \end{subfigure}
        \begin{subfigure}[b]{0.23\textwidth}  
            \centering 
            \includegraphics[width=\textwidth]{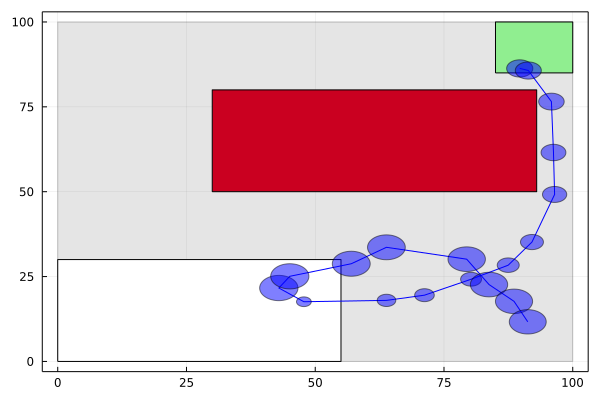}
            \caption[]%
            {{\footnotesize 5 second planning time}}    
            \label{fig:replan2}
        \end{subfigure}
        \caption[ Runtime results of belief-SST with Wasserstein metric and $p_{bias} =0.02$.]
        {\small Runtime results of belief-SST with Wasserstein metric and $p_{bias} =0.02$, with $2\sigma$ Gaussian covariance ellipses.} 
        \label{fig:unicycle}
        \vspace{-3.75mm}
    \end{figure}

Fig.~\ref{fig:unicycle} shows an example of motion plans computed by belief-SST for Environment 3 with the feedback linearized unicycle. Given a short planning time limit, belief-SST quickly finds an initial solution through the larger passageway. As planning time increases, belief-SST improves the solution, converging to one with lower cost by visiting the measurement region before traversing the small passage.

\subsection{Online planning demonstration}
\begin{figure}
        \centering
        \begin{subfigure}[b]{0.23\textwidth}
            \centering
            \includegraphics[width=\textwidth]{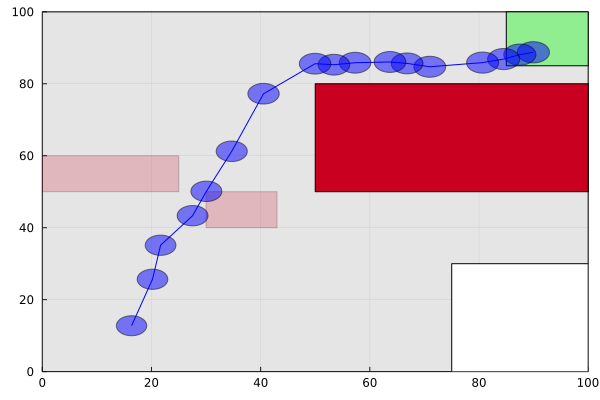}
            \caption[Initial plan]%
            {{\footnotesize Initial Plan}}    
            \label{fig:initial3}
        \end{subfigure}
        \begin{subfigure}[b]{0.23\textwidth}  
            \centering 
            \includegraphics[width=\textwidth]{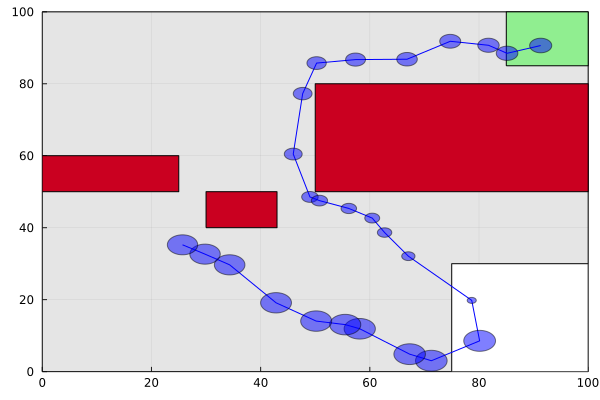}
            \caption[]%
            {{\footnotesize 0.5 second re-plan}}    
            \label{fig:replan3}
        \end{subfigure}
        \caption[  ]
        {\small Online planning scenario. The robot is initially unaware of the obstacles on the left, and plans according to the red obstacle on the right (a). After updating the environment, the robot finds a safe path by localizing at the measurement region in white (b).} 
        \label{fig:replanning}
        \vspace{-5mm}
    \end{figure}

Due to the improved speed of computation, our algorithm is able to plan in an online fashion. Fig.~\ref{fig:replanning} shows a simple example showcasing the algorithm's effectiveness in re-planning online when changes to the problem such as new obstacles or measurements are detected. In this example, the robot initially plans with the knowledge that there is one obstacle in red, and finds a motion plan that reaches the goal region. Note that in this plan, there is no need to visit the measurement region since the chance constraint can be satisfied without doing so. However, as the robot moves, it updates the environment after detecting obstacles and re-plans. In this updated environment, moving directly to the goal will not satisfy the chance constraint. Therefore, the robot finds a solution plan that moves to the measurement region to localize before traversing safely to the goal region.

%% file: sections/Conclusion.tex
\section{Conclusion}
This paper proposes belief-$\mathcal{A}$, a general framework for converting any kinodynamical sampling-based tree planner to the belief space for linear systems. The proposed method preserves the convergence properties of the underlying algorithm, and empirical tests demonstrate its efficacy. Future work includes implementing the framework on an actual robotic platform and extending the method to work with a broader class of motion planning algorithms and belief distributions.

%% file: sections/Appendix.tex
\include{sections/extended/benchmarktable}

%% file: sections/extended/benchmarktable.tex
\appendix
\subsection{Benchmarking Results}
\label{appendix:benchmarks}
\begin{sidewaystable}
    \centering
    \footnotesize
    \begin{tabular}{|c|c|c|c|c|c|c|c|c|c|}
    \hline
    & \multicolumn{3}{c|}{Environment 1} & \multicolumn{3}{c|}{Environment 2} & \multicolumn{3}{c|}{Environment 3}\\
    \hline
    Algorithm & Time (1st) & cost (1st) & 
    Cost (10 s) & Time (1st) & Cost (1st) & 
    Cost (10 s) & Time (1st) & Cost (1st) & 
    Cost (10 s)\\
    \hline
     \multicolumn{10}{|c|}{2D System}\\
    \hline
     RRBT   & $0.17 \pm 0.016$ & $\mathbf{113.8 \pm 0.23}$ & $\mathbf{109.8 \pm 0.032}$ &  1.56 $\pm 0.116$ & $\mathbf{216.08 \pm 0.811}$ & $\mathbf{199.17 \pm 0.626}$ & 2.24 $\pm 0.11$ & $\mathbf{146.8 \pm 0.221}$ & $\mathbf{121.2 \pm 0.437}$\\\hline
     B-RRT ($l_2$)& $\mathbf{0.003 \pm 0.0002}$ & $194.8 \pm 3.91$  & 192.6 $\pm 3.51$ & $\mathbf{0.024 \pm 0.002}$ & 360.75 $\pm 4.57$ & 341.72 $\pm 4.336$ & $\mathbf{0.009 \pm 0.0008}$ & 293.04 $\pm 4.76$ & 292.61 $\pm 3.76$\\\hline
     B-RRT ($W_2$)&  $0.029 \pm 0..053$ & 195.7 $\pm 3.60$ & $186.1 \pm 3.62$ & $0.2 \pm 0.011$ & 353.7 $\pm 5.66$ & 345.1 $\pm 3.96$ & $0.016 \pm 0.0009$ & 288.6 $\pm 3.65$ & $285.3 \pm 4.09$ \\\hline
     B-RRT ($W_2$, $0.2\,p_{bias}$) & $0.029 \pm 0.006$ & $189.8 \pm 4.09$ & $185.8 \pm 3.05$ & $0.071 \pm 0.0036$  & 348.4 $\pm 4.05$ & 340.0 $\pm 4.27$ & 0.016 $\pm 0.0008$ & 294.93 $\pm 3.68$ & 293.81 $\pm 3.51$ \\\hline
     B-SST ($l_2$) & $\mathbf{0.004 \pm 0.0003}$ & $163.6 \pm 3.24$ & $\mathbf{111.8 \pm 0.124}$ & $2.62 \pm 0.38$ & 259.4 $\pm 2.95$ & 236.09 $\pm 2.83$ & $\mathbf{0.008 \pm 0.0008}$ & 264.94 $\pm 4.82$ & 166.85 $\pm 0.88$\\\hline
     B-SST ($W_2$) & $0.029 \pm 0.0048$ & $181.3 \pm 3.83$  & $\mathbf{111.1 \pm 0.13}$ & $0.51 \pm 0.05$ & $295.2 \pm 4.17$ & $236 \pm 2.79$ &  $0.043 \pm 0.0038$ & $263.9 \pm 2.94$  & $157.4 \pm 1.56$ \\\hline
     B-SST ($W_2$, $0.2\, p_{bias}$) & 0.027 $\pm 0.003$ & 168 $\pm 3.23$ & $\mathbf{110.2 \pm 0.143}$ & 0.152 $\pm 0.01$& 304.72 $\pm$ 3.76 & $\mathbf{204.87 \pm 0.833}$ &0.041 $\pm$ 0.0036 & 263.16 $\pm 3.42$ & $\mathbf{128.62 \pm 0.631}$ \\\hline
      \multicolumn{10}{|c|}{Linearized Unicycle}\\
     \hline
     RRBT    & $0.225 \pm 0.019$  & $\mathbf{123.9 \pm 0.218}$ & $\mathbf{120 \pm 0.042}$ & $15.4 \pm 1.08$ & $\mathbf{227.1 \pm 0.84}$ & $\mathbf{222.6 \pm 0.34}$ & $8.95 \pm 0.563$ & $\mathbf{201.6 \pm 1.88}$ & $199.4 \pm 1.73$ \\\hline
     
     B-RRT ($l_2$)& $\mathbf{0.011 \pm 0.001}$ & $250 \pm 6.64$ & $250 \pm 7.22$ & $\mathbf{0.447 \pm 0.0334}$ & $399.4 \pm 7.22$ &$367.7 \pm 5.83$  & $0.045 \pm 0.003$ & $391.9 \pm 8.21$ & $390.9 \pm 8.1$ \\\hline
     
     B-RRT ($W_2$)& $0.034 \pm 0.004$  & $247.6 \pm 4.13$ & $221 \pm 3.88$ & $0.577 \pm 0.0468$ & $286.4 \pm 7.51$ & $256.3 \pm 6.85$ & $0.2 \pm 0.015$ & $341.8 \pm 6.23$&$268.3 \pm 9.9$ \\\hline
     
     B-RRT ($W_2$, $0.2\,p_{bias}$) & $0.041 \pm 0.004$ & $244.8 \pm 3.49$  & $231.6 \pm 4.21$ & $0.61 \pm 0.048$  & $287 \pm 8.93$ & $260.8 \pm 7.73$ & $0.19 \pm 0.017$ & $332.5 \pm 6.64$ & $289.3 \pm 8.46$\\\hline
     
     B-SST ($l_2$) & $0.013 \pm 0.001$ & $198.7 \pm 4.44$ & $138.5 \pm 1.46$ & $7.08 \pm 0.69$ & $267.4 \pm 3.02$& $\mathbf{237.9 \pm 3.02}$ & $\mathbf{0.035 \pm 0.0037}$ & $339.3 \pm 6.62$& $\mathbf{177.8 \pm 5.09}$ \\\hline
     
     B-SST ($W_2$) & $0.055 \pm 0.0048$ & $190.3 \pm 4.14$ & $137.2 \pm 1.27$ & $2.19 \pm 0.242$ & $277.8 \pm 4.02$ & $\mathbf{232.1 \pm 2.07}$ & $0.193 \pm 0.014$ & $347.2 \pm 6.2$ & $188.7 \pm 3.5$\\\hline
     
     B-SST ($W_2$, $0.2\, p_{bias}$) & $0.054 \pm 0.0053$ & $193.8 \pm 4.52$ & $140.4 \pm 1.52$  &  $0.88 \pm 0.11$ & $274.5 \pm 4.60$ & $\mathbf{231.1 \pm 2.44}$ & $0.153 \pm 0.0148$ & $316.1 \pm 5.81$ & $\mathbf{164.4 \pm 1.52}$\\\hline
    \end{tabular}
    \caption{\small Benchmark planner performance results. Each entry is the mean of 100 simulations, with standard error of the mean bounds. The scores within 10\% of the best score in each column are bolded.}
\end{sidewaystable}